\documentclass[preprint,12pt,authoryear]{elsarticle}

\usepackage{xcolor}

\usepackage{amssymb}
\usepackage{amsmath}
\usepackage{graphicx}
\usepackage{booktabs}
\usepackage{hyperref}
\usepackage{multirow}
\usepackage{pifont}
\newcommand{\cmark}{\ding{51}} % check mark
\newcommand{\xmark}{\ding{55}} % x mark

\usepackage{adjustbox}

\journal{Nuclear Physics B}

\begin{document}

\begin{frontmatter}



\title{ProfVLM: A lightweight video-language model for multi-view proficiency estimation} %% Article title

\author[label1]{Edoardo Bianchi}
\author[label2]{Jacopo Staiano}
\author[label1]{Antonio Liotta}

\affiliation[label1]{organization={Free University of Bozen-Bolzano},
             addressline={Via Bruno Buozzi 1},
             city={Bozen-Bolzano},
             postcode={39100},
             country={Italy}}

\affiliation[label2]{organization={University of Trento},
             addressline={Via Inama 5},
             city={Trento},
             postcode={38122},
             country={Italy}}

%%\author{} %% Author name

%%\affiliation{organization={},%Department and Organization
            %%addressline={}, 
            %%city={},
            %%postcode={}, 
            %%state={},
            %%ountry={}}

%% Abstract
\begin{abstract}
Most existing approaches formulate action quality assessment and skill proficiency estimation as discriminative prediction tasks, typically producing discrete labels or scores without explicitly modeling the reasoning process underlying the assessment. We instead reformulate the problem as generative vision-language modeling, introducing ProfVLM, a parameter-efficient vision-language model that jointly predicts proficiency levels and generates expert-like natural language feedback from multi-view videos.
% Unlike classification-based methods, 
% that operate as black boxes, 
ProfVLM leverages conditional language generation to provide actionable insights along with quantitative evaluation scores. Central to our method is an AttentiveGatedProjector that dynamically fuses and projects multi-view egocentric and exocentric features from a frozen TimeSformer backbone into a language model fine-tuned for feedback generation. Trained on EgoExo4D with expert commentaries, ProfVLM surpasses state-of-the-art methods while using up to 20x fewer parameters and reducing training time by up to 60\% compared to existing classification-based methods. By providing natural language critiques aligned with performance levels, this work shows that generative vision-language modeling offers a powerful and efficient paradigm shift for interpretable action quality assessment.
\end{abstract}



%% Keywords

\end{frontmatter}


\section{Introduction}
Understanding proficiency in complex activities from video is crucial for applications such as coaching and rehabilitation. Unlike action recognition, proficiency estimation and Action Quality Assessment (AQA) require a deep and technical understanding of the actions, a non-trivial task even for state-of-the-art architectures.

Moreover, prior approaches present limitations that restrict their practical application. First, proficiency estimation is commonly formulated as a discriminative prediction task (classification or regression), relying on dedicated output heads that produce scores without explicitly modeling the reasoning process experts use when assessing performance. Second, most prior works focus on single-view analysis, despite comprehensive skill assessment often requiring multi-camera perspectives to capture complementary aspects such as hand positioning from egocentric views and overall technique from exocentric angles. Third, although language signals have been incorporated in some approaches, they are typically used as auxiliary supervision or in parallel tasks, rather than as a unified reasoning mechanism that directly drives proficiency prediction. As a result, prediction and explanation are optimized through separate objectives, limiting the model’s ability to align visual evidence with linguistic reasoning in a unified manner.

We address these gaps by reframing proficiency estimation as a conditional generation problem. This formulation enables the model to predict proficiency levels while producing expert-style explanatory commentary, mirroring how human evaluators justify their assessments. Building on this idea, we develop ProfVLM, a parameter- and training-efficient vision-language model that jointly outputs proficiency labels and interpretable, expert-like feedback from multi-view video inputs.

This paper introduces the following key contributions:

\begin{itemize}
\item A paradigm shift from classification to a generative formulation of proficiency estimation, enabling models to produce both proficiency labels and natural language commentary;
\item ProfVLM, a parameter- and training-efficient vision-language model for generative proficiency estimation that handles multi-view inputs and generalizes across multiple domains;
\item An AttentiveGatedProjector module that performs structured multi-view fusion through cross-view attention and learnable gating, aligning video representations with the LLM embedding space.
\end{itemize}

To the best of our knowledge, ProfVLM is the first vision-language model to perform multi-view proficiency estimation entirely within an autoregressive generative framework, without relying on a dedicated classification head. The results show how the proposed ProfVLM improves over baselines in multi-view accuracy, while using 20x fewer parameters (5.3M vs. 121M), 2x fewer frames (8 vs. 16), and 60\% reduction in training time (6 vs. 15 epochs). Compared to recently published works, our model again achieves better accuracy using 5x fewer parameters (5.3M vs. 27M) and 2-4x fewer frames (8 vs. 16/24/32). This efficiency highlights the benefits of integrating language into skill assessment from multiple visual viewpoints. Additionally, ProfVLM introduces high-quality feedback generation (BERTScore F1 of 85.53), a feature not present in previous approaches, which provides actionable insights for users.

\begin{figure}[t]
  \centering
  \includegraphics[width=1\textwidth]{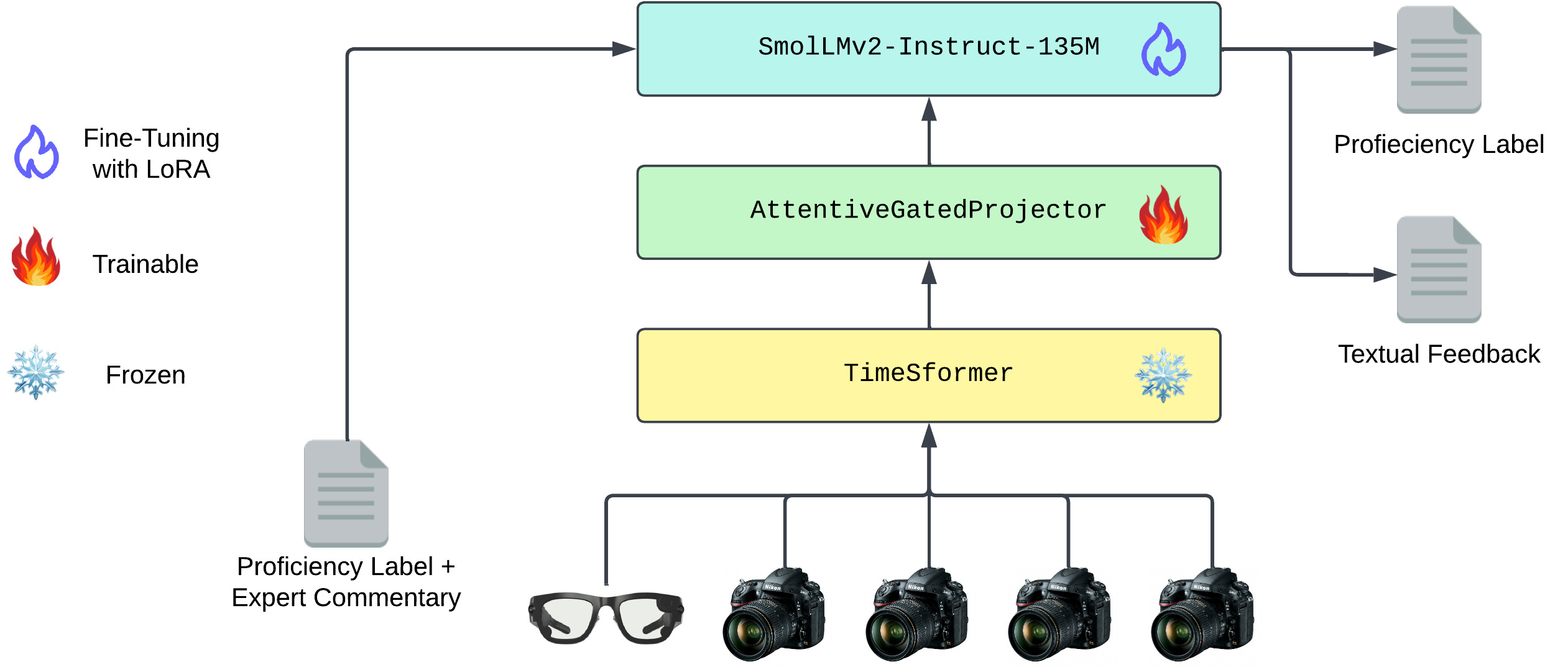}
  \caption{ProfVLM architecture. Multi-view video is encoded by a frozen TimeSformer, fused via a trainable AttentiveGatedProjector, and decoded by a LoRA-tuned SmolLM2 to generate both proficiency label and commentary.}
  \label{fig:model}
\end{figure}

\section{Background and Related Works}
The intersection of computer vision and sports analysis continues to expand across multiple domains, from team sports officiating systems that leverage multi-view analysis for foul detection \citep{football_var} to broadcast enhancement and performance analytics \citep{sports_cv_overview, sport_cv_2}. These applications demonstrate the growing integration of AI technologies in sports contexts, highlighting the need for robust and interpretable assessment systems that can operate across different skill levels and sporting disciplines.

Action recognition remains a highly active and evolving research area, with comprehensive surveys highlighting its rapid progress across diverse domains and methodologies \citep{actionrecreview, actionrecreview1}. Recent advances have explored temporal action segmentation in sparse action scenarios \citep{figure_skating}, multi-scale feature fusion with video-text constraints \citep{unimultnet}, graph neural networks for child activity recognition \citep{child_car}, and skeleton-enhanced spatio-temporal networks for sport action recognition \citep{gsp}. Domain-specific applications have demonstrated the importance of tailored tracking solutions, as shown in skiing performance analysis where specialized tracking algorithms have been developed to handle the unique challenges of winter sports environments \citep{skier_tracking}.

Moving beyond classical action recognition tasks, Action Quality Assessment (AQA) and Proficiency Estimation target the evaluation of how well an action is performed. Early deep learning approaches relied on video-only inputs and direct score regression using 3D CNNs—see e.g. \citet[C3D]{c3d}, and \citet[I3D]{carreira17}—or transformer-based models—see e.g., \citet[VST]{videoswin}—often optimized for continuous or discrete skill labels. More recent approaches employ spatio-temporal feature extraction and deep learning or self-supervised and transfer learning for action assessment in rehabilitation \citep{hiprehab, sslrehab}. Although effective, these models offer limited interpretability and require domain-specific supervision \citep{AQA_survey}.

Multi-modal AQA and Proficiency Estimation enhance robustness and expressiveness by leveraging complementary modalities such as RGB, optical flow, depth, pose, and audio. Audio-visual fusion models such as Skating-Mixer \citep{skatinmixer} adopt hierarchical integration strategies to exploit temporal alignment and rhythm. Others use RGB-depth pairs for industrial tasks \citep{gsfmeccano} or combine skeletal motion with visual input to capture nuanced kinematics in sport performance \citep{gsp, hierarchicalpose, skeletonAR}. Recent work has also explored domain-specific pose estimation challenges, such as extending human skeleton estimation to include particular sports equipment \citep{ski_pose_estimation}. Recent benchmarks such as EgoExoLearn \citep{egoexolearn} and EgoExo4D \citep{egoexo4d} underscore the value of multi-view egocentric and exocentric perspectives. However, the baselines proposed in these works demonstrate that the task is far from trivial, emphasizing the critical need for specialized architectures and tailored training objectives. Parameter-efficient architectures like SkillFormer \citep{skillformer} have further advanced multi-view proficiency estimation by using cross-attention mechanisms, while physiological signal extraction approaches such as EgoPulseFormer \citep{egoppg} demonstrate how incorporating heart rate data from egocentric cameras can significantly improve skill assessment. Recent advances in temporal sampling strategies have demonstrated that effective skill assessment requires domain-specific adaptations, as different sports and activity categories require distinct temporal parameters to capture fundamental movement patterns while preserving temporal continuity \citep{pats}. While prior works incorporate language components or narrative evaluation \citep{parmar2019mtl, zhang2024nae}, they maintain separate discriminative heads for score or label prediction. In contrast, our approach embeds proficiency prediction directly within the generative language modeling objective, unifying classification and explanation into a single autoregressive process.

In parallel, Vision-Language Models (VLMs) have been extended from image-text to video-text alignment. VideoBERT \citep{videobert} pioneered joint video-text modeling by learning bidirectional joint distributions over sequences of visual and linguistic tokens derived from vector quantization. Image-focused VLMs like LLaVA \citep{llava} established strong baselines for visual instruction tuning and MiniGPT-4 \citep{minigpt4} demonstrated effective vision-language alignment with frozen image encoders. VideoLLaMA \citep{videollama} introduced instruction-tuned audio-visual language modeling with dual-branch architecture for comprehensive video understanding, while LLaMA-VID \citep{llamavid} focused on efficient video comprehension through temporal token compression. Video-ChatGPT \citep{videogpt} and Video-LLaVA \citep{videolava} further unified image and video modalities but initially lacked robust temporal grounding. Recent approaches include VALOR \citep{valor}, which improved video-text alignment through contrastive learning, and SmolVLM \citep{smolvlm}, a compact model that achieves state-of-the-art performance for its memory footprint while maintaining competitive multimodal capabilities. Advanced models such as Video-STaR \citep{videostar} introduced instruction-tuning with labeled video corpora, while VideoAgent \citep{videoagent} and Apollo \citep{apollo} framed video understanding as decision-making, improving long-form comprehension with efficient token compression, curriculum learning, or summarization tokens.

Computational efficiency remains a key consideration in both image- and video-based analysis, with recent advances combining knowledge distillation with tensor decomposition \citep{kdohsvd} and applying pruning to transformer architectures to maintain accuracy while reducing computational costs \citep{pose_pruning}. Such efficiency considerations are particularly important for real-time applications in sports analysis and skill assessment, where rapid feedback can be crucial for training and performance improvement.

Our proposed ProfVLM bridges generative VLMs and skill assessment by treating proficiency estimation as a conditional language generation task. Built on frozen video transformers and LoRA-tuned language models, it jointly predicts structured proficiency levels and generates expert-like feedback.

\section{Proposed Methodology}
We introduces ProfVLM, a multimodal vision-language model for skill assessment from videos captured from multiple perspectives. The architecture comprises a pre-trained TimeSformer \citep{timesformer} for visual encoding, an AttentiveGatedProjector
% —extending the approach from \citealp{skillformer}—
which fuses multi-view features and projects them into the language embedding space, and a parameter-efficient adaptation of the open-source SmolLM2-135M-Instruct \citep{smollm2}, fine-tuned using LoRA \citep{lora}. Figure~\ref{fig:model} depicts the complete framework.

\subsection{Problem Formulation}
The task of demonstrator proficiency estimation involves analyzing multi-view video recordings of a subject performing a task, with the goal of assessing their skill level and providing qualitative feedback. Each sample consists of time-synchronized video streams from multiple viewpoints, specifically including one egocentric (first-person) and up to four exocentric (third-person) perspectives. Given this input, the objective is twofold: (1) predicting the demonstrator's proficiency level as one of four predefined categories (Novice, Early Expert, Intermediate Expert, or Late Expert), and (2) generating a natural language comment describing their performance. 

Formally, for a given set of $n$ synchronized video streams ${V_1, V_2, …, V_n}$, the model must produce a textual feedback $c \in \mathcal{T}$, where $\mathcal{T}$ denotes the space of valid commentary. The discrete proficiency label $y \in \{0, 1, 2, 3\}$ is implicitly generated as part of this feedback rather than being predicted through a dedicated classification head.

\subsection{Video Feature Extraction}
\label{subsec:feature_extraction}
We employ a TimeSformer backbone \citep{timesformer}, pretrained on Kinetics-600 \citep{carreira17}, to extract spatio-temporal representations from video. Our choice of TimeSformer is motivated by two key factors: first, it serves as the established baseline architecture for evaluation in the EgoExo4D dataset \citep{egoexo4d}, enabling direct comparison with existing benchmarks; second, its divided space-time attention mechanism is specifically designed for video understanding, making it inherently superior to image-focused architectures that lack temporal modeling capabilities for proficiency estimation tasks. We retain the model frozen throughout training, without any fine-tuning or adaptation.

For multi-view scenarios, we employ parallel processing by flattening the batch and view axes, enabling simultaneous feature extraction across all synchronized video streams through our shared frozen backbone. The extracted features are subsequently reshaped to maintain distinct view-specific representations before entering the fusion and projection module, preserving the view separation critical for effective cross-view integration.

Inputs are fixed to 8-frame clips, matching the TimeSformer pretraining configuration. Preserving this format ensures compatibility with pretrained temporal attention patterns and avoids introducing additional trainable parameters, aligning with our efficiency objectives. Moreover, increasing frames yields diminishing returns: the official TimeSformer reports only a 3.9\% Acc@1 gain when moving from 8 to 96 frames on Kinetics-600 despite a 12$\times$ increase in input size\footnote{Results from official implementation: \url{https://github.com/facebookresearch/TimeSformer}}. Despite using fewer frames, our architecture achieves state-of-the-art performance compared to baselines with 16-32 frames, leveraging more effective feature utilization and synchronized multi-view sampling to maintain consistent semantics across perspectives.

\begin{figure}[t]
  \centering
  \includegraphics[width=0.6\textwidth]{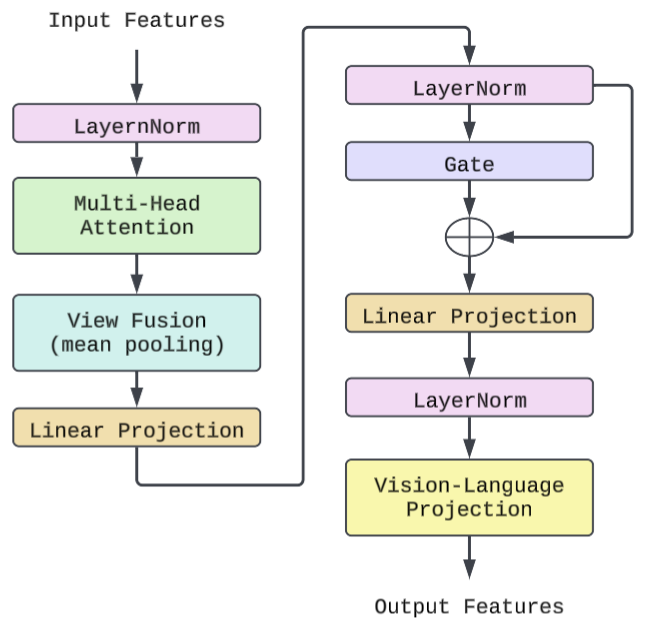}
  \caption{AttentiveGatedProjector architecture. Multi-view input features are initially normalized and combined using multi-head attention. The aggregated representation is then refined through a feed-forward block with residual paths and a trainable gating unit. Finally, the output undergoes linear projection, normalization, and alignment to the language model space via learned scaling and shifting parameters.}
  \label{fig:AttentiveGatedProjector}
\end{figure}

\subsection{AttentiveGatedProjector: Multi-View Vision-Language Alignment}
\label{subsec:attentive_projector}
We design the AttentiveGatedProjector, an architecture tailored for integrating multi-view video representations and aligning them with the input space of large language models. Though bearing resemblance to a concurrent work \citep{skillformer} which targets classification, our module is instead optimized for joint vision-language modeling. While both approaches employ attention-based fusion mechanisms, they operate in fundamentally different learning paradigms: SkillFormer implements a discriminative classification framework that maps multi-view videos through feature fusion and classification heads to discrete labels, whereas our AttentiveGatedProjector enables a generative vision-language framework that projects multi-view videos through vision-language alignment into language models for natural language explanations with embedded labels.

This difference necessitates distinct architectural components and optimization objectives. Unlike discriminative fusion modules that optimize for classification, our projector is designed for video-language integration and includes statistical distribution matching between visual and textual embeddings through learnable normalization parameters, cross-modal dimensional projection to match the target language model's embedding space, and preservation of semantic coherence across autoregressive generation sequences—requirements that are absent in classification-only architectures. The full architecture is illustrated in Figure~\ref{fig:AttentiveGatedProjector} and detailed below.

\textbf{Independent View Normalization:}
Layer normalization is applied to each view (e.g., ego- or exo-centric) independently. This ensures that statistical disparities across views are normalized early, reducing modality-specific biases before fusion.

\textbf{Cross-View Multi-Head Attention:}
To learn inter-view dependencies, we employ a multi-head attention mechanism where each view can contextualize its representation based on others through self-attention operations. This step flexibly captures complementary cues among perspectives, with attention weights dynamically determining the relevance of each view for the current sample.

\textbf{Learnable Gating Mechanism}:
The fused features are passed through a gated transformation layer comprising a feed-forward network followed by a learnable sigmoid gate. This gate modulates each hidden unit independently through element-wise multiplication, enhancing discriminative signals while suppressing irrelevant or redundant information. The gating mechanism is particularly crucial for generative tasks where noise suppression affects text coherence.

\textbf{Vision-Language Projection:}
Finally, the fused video features undergo a two-stage projection process specifically designed for vision-language integration. The features are first mapped to the target language model dimensionality, then undergo statistical normalization that centers them to zero mean, scales to unit variance, and applies learnable parameters to match the statistical distribution of language model token embeddings. This statistical alignment is essential for maintaining semantic coherence during text generation.

\begin{table}[t]
\centering
\begin{adjustbox}{width=\textwidth}
\begin{tabular}{@{}p{2cm}p{13.5cm}@{}}
\toprule
\textbf{Role} & \textbf{Content} \\
\midrule
System & \texttt{You are a visual agent for human performance analysis.} \\
User & \texttt{Here are 8 frames sampled from a video: <|video\_start|><|video|><|video\_end|>. Given this video, analyze the proficiency level of the subject.} \\
Assistant & \texttt{Proficiency Level: Intermediate Expert.
Proficiency Commentary: The subject demonstrates smooth and controlled movements throughout the sequence, with clear intent and minimal hesitation.} \\
\bottomrule
\end{tabular}
\end{adjustbox}
\caption{Prompt format used for training ProfVLM. During inference, only the System and User messages are provided, and the model generates both the proficiency level classification and explanatory commentary in a unified text response.}
\label{tab:prompt_format}
\end{table}

\subsection{Tokenization and LLM Prompt Structure}
\label{sec:tokenization}
To enable conditional generation, we extend the tokenizer vocabulary with three special tokens: \texttt{<|video\_start|>}, \texttt{<|video\_end|>}, and \texttt{<|video|>}. These delimiters explicitly mark the location of the visual input and allow the language model to attend to the fused video features in a controlled way. Importantly, our model performs both tasks—proficiency classification and commentary generation—entirely through language modeling. That is, the discrete proficiency label is not produced via a separate classification head but is instead embedded as part of the autoregressively generated textual response.

We adopt a structured prompt composed of system, user, and assistant messages in a chat-style format, as shown in Table~\ref{tab:prompt_format}. Our approach follows the chat template used during the pretraining and instruction-tuning of the open-source SmolLM2-135M-Instruct \citep{smollm2} to maintain consistency with the model's expected input structure. The system message defines the model’s role; the user message includes the visual input placeholder bracketed by the special tokens; and the assistant message contains the target label and commentary. During training, all three messages are used to condition and supervise generation. During inference, only the system and user messages are provided as input, and the model is required to generate the assistant message, producing both the predicted proficiency label and the natural language feedback as a textual output.

The structured output format is provided by the training instance targets, allowing the model to progressively learn the required output schema during supervised fine-tuning. Since the inference prompt does not explicitly enforce the template, schema compliance emerges during fine-tuning: the ablations reported in Sec.~\ref{sec:ablation_format} show that before fine-tuning the model does not follow the required output format, but after only a single epoch it achieves full compliance for the label field and near-complete compliance for the commentary field. At convergence, both fields reach 100\% compliance, indicating that the autoregressive decoder reliably learns to generate template-aligned outputs under structured supervision. For deployment scenarios requiring strict guarantees, constrained decoding strategies\footnote{Outlines: \url{https://github.com/dottxt-ai/outlines}} can be applied to enforce schema compliance.

\begin{table}[t]
\centering
\begin{tabular}{lcccc}
\toprule
& \multicolumn{2}{c}{\textbf{Official}} & \multicolumn{2}{c}{\textbf{Experiment}} \\
\cmidrule(lr){2-3} \cmidrule(lr){4-5}
\textbf{Scenario} & Train & Val & Train & Val \\
\midrule
Basketball & 576 & 105 & 1104 & 105 \\
Cooking & 83 & 20 & 381 & 20 \\
Dancing & 408 & 124 & 1220 & 124 \\
Music & 149 & 39 & 431 & 39 \\
Bouldering & 620 & 162 & 923 & 162 \\
Soccer & 68 & 16 & 179 & 16 \\
\midrule
Total & 1904 & 466 & 4238 & 466 \\
\bottomrule
\end{tabular}
\caption{Comparison of original and experimental dataset statistics for train and validation splits. Our train set is larger because each sample includes multiple expert commentary annotations. The validation set is kept unchanged to ensure a fair comparison.}
\label{tab:dataset-statistics}
\end{table}

\section{Experimental Setup}
\subsection{Dataset}
We conduct experiments on the EgoExo4D dataset \citep{egoexo4d}, which contains over 1,200 hours of aligned egocentric and exocentric video from 740 participants across 123 locations. The dataset spans eight activities—cooking, health, bike maintenance, music, basketball, bouldering, soccer, and dance—captured in diverse real-world settings.

For the demonstrator proficiency estimation benchmark, six domains are used: cooking, music, basketball, bouldering, soccer, and dance. Each sample consists of one first-person video (Project Aria glasses \citep{projectaria}) and up to four third-person views (GoPro cameras), alongside synchronized audio, eye gaze, 3D pose, and language annotations. Notably, the proficiency label distribution is skewed toward higher skill levels (Intermediate and Late Experts), reflecting the dataset's intentional focus on recruiting participants with advanced abilities.

\begin{figure}[t]
  \centering
  \includegraphics[width=1\textwidth]{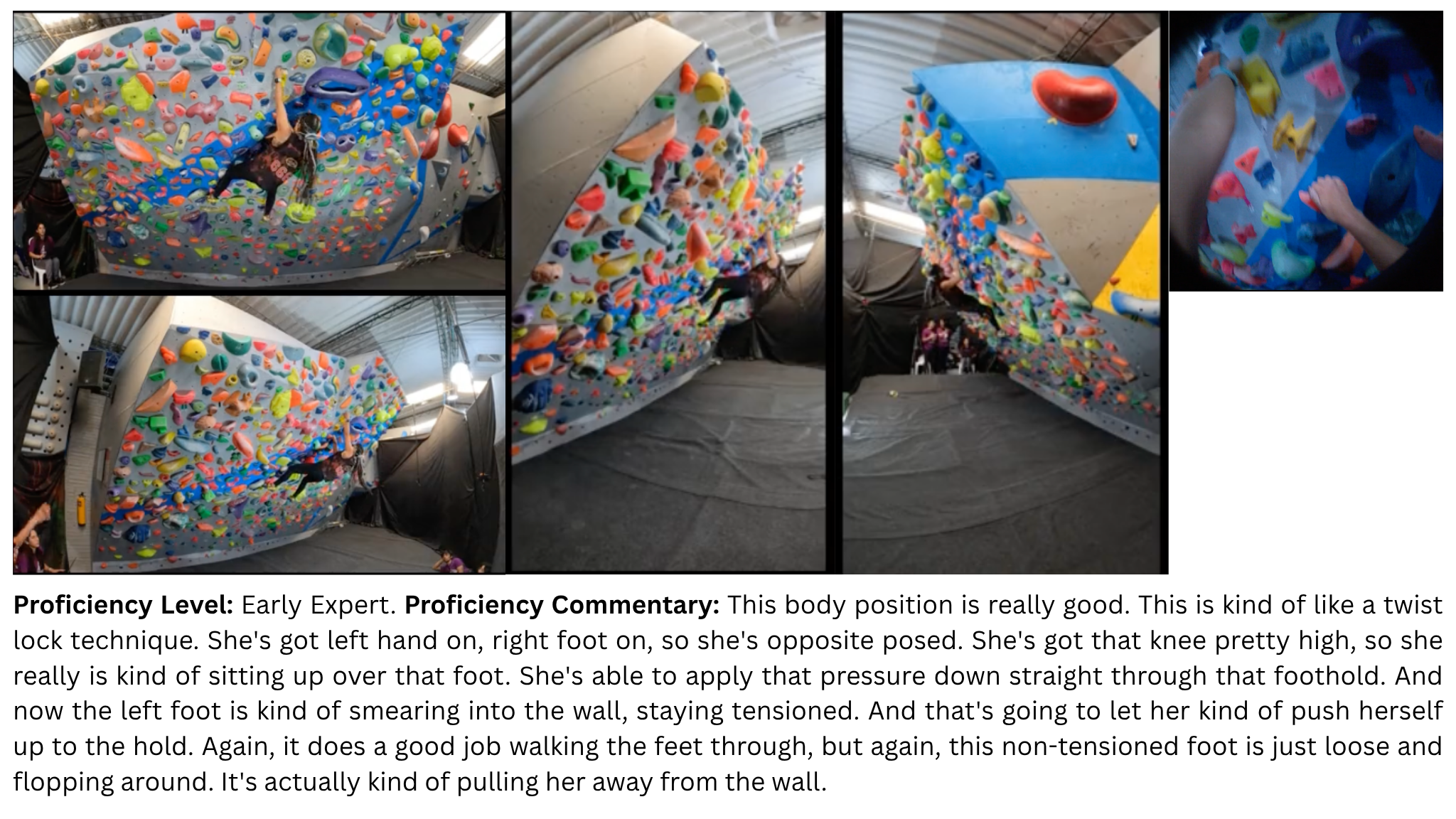}
  \caption{Synchronized frames from the five views (1 egocentric and 4 exocentric) of a single EgoExo4D test sample for the bouldering domain. The proficiency level and commentary apply to the entire multi-view sequence, demonstrating the multi-perspective skill assessment setup.}
  \label{fig:dataExample}
\end{figure}

The dataset provides two key supervision signals (Fig. \ref{fig:dataExample}): a discrete proficiency level (\textit{Novice}, \textit{Early Expert}, \textit{Intermediate Expert}, or \textit{Late Expert}) and a free-form expert commentary from domain specialists. These commentaries were originally provided as verbal feedback by domain experts and subsequently transcribed from raw audio without manual curation or editing. This approach results in commentary text that preserves the evaluation style of domain experts, including spontaneous observations, colloquial language, and the natural flow of expert reasoning. However, it also introduces potential challenges such as transcription errors, incomplete sentences, repetitions, and varying levels of detail across commentaries.

We use all videos with proficiency estimation labels from the dataset, adhering to the official benchmark's training and validation splits. As in prior work \citep{egoppg}, we train on the official training set and evaluate on the full, held-out validation set to ensure comparability. 

Our training set contains more samples than the official benchmark because each unique video-commentary pair is treated as a separate training instance. Classification labels are explicitly integrated into the input text as prompts (see Sec.~\ref{sec:tokenization} and Tab.~\ref{tab:prompt_format}), allowing the model to leverage textual and label information jointly. This provides richer supervision compared to classification baselines that rely solely on discrete labels, but also introduces a more challenging generative objective: the model must produce consistent proficiency predictions across varied linguistic contexts rather than mapping to fixed label embeddings. Table~\ref{tab:dataset-statistics} summarizes the dataset statistics, highlighting the expanded training size while preserving the original validation set for fair evaluation.

\subsection{Implementation Details}
\label{sec:implementationDet}
We fine-tune ProfVLM in three view configurations: Ego (1 egocentric view), Exos (4 exocentric views), and Ego+Exos (all 5 views combined). Our training protocol consists of 6 epochs with a batch size of 32, using a cosine annealing learning rate schedule starting at 3e-4 with 1 warmup epoch. All experiments run on a single NVIDIA A100 GPU (80GB).

For video processing, we employ a frozen TimeSformer \citep{timesformer} pre-trained on Kinetics-600 \citep{carreira17}. From each video, we extract 8 uniformly distributed frames from a randomly positioned temporal window, applying a sampling rate of 8 to ensure temporal coverage across variable-length clips. Each frame is pre-processed by resizing its shortest edge to 224 pixels, center-cropping to $224 \times 224$, rescaling to $[0, 1]$, and normalizing with mean $[0.45, 0.45, 0.45]$ and standard deviation $[0.225, 0.225, 0.225]$.

For language modeling, we use the open-source SmolLM2-135M-Instruct \citep{smollm2}, fine-tuned via Low-Rank Adaptation (LoRA) \citep{lora} with rank $r = 8$ and scaling factor $\alpha = 32$. The AttentiveGatedProjector variant implements a hidden size of 1024 with 4 attention heads, while the simpler MLP variant uses a hidden size of 576 to match the LLM's input dimension.

To further validate the contributions of our approach, we conduct ablation studies (Sec.~\ref{sec:ablations}) examining (1) architectural components, (2) supervision modality and volume, and (3) prompt design choices.

\subsection{Evaluation Metrics}
We employ a comprehensive evaluation framework that addresses both the classification and natural language generation aspects of our model. For classification, we use accuracy as the primary metric since it is the standard measure in the official EgoExo4D benchmark \citep{egoexo4d}, ensuring direct comparability with existing methods. Given the class imbalance present in the dataset, which can affect accuracy scores, we additionally report F1-score to provide a more robust evaluation.

\begin{table}[t]
\centering
% \small
\begin{adjustbox}{width=\textwidth}
\begin{tabular}{lccccccc}
\toprule
\textbf{Method} & \textbf{Pretrain} & \textbf{Ego} & \textbf{Exos} & \textbf{Ego+Exos} & \textbf{Frames} & \textbf{Train Params} & \textbf{Epochs} \\
\midrule
Random & - & 24.9 & 24.9 & 24.9 & - & - & - \\
Majority-class & - & 31.1 & 31.1 & 31.1 & - & - & - \\
TimeSFormer & - & 42.3 & 40.1 & 40.8 & 16 & 121M & 15 \\
TimeSFormer & K400 & 42.9 & 39.1 & 38.6 & 16 & 121M & 15 \\
TimeSFormer & HowTo100M & \textbf{46.8} & 38.2 & 39.7 & 16 & 121M & 15 \\
TimeSFormer & EgoVLP & 44.4 & 40.6 & 39.5 & 16 & 121M & 15 \\
TimeSFormer & EgoVLPv2 & \underline{45.9} & 38.0 & 37.8 & 16 & 121M & 15 \\
\midrule
PandaGPT (0-shot) & ImageBind + Vicuna & 23.5 & 25.6 & 25.8 & - & 52M & - \\
SmolVLM (0-shot) & - & 34.2 & 30.7 & 31.1 & - & 256M & - \\
SmolVLM2 (0-shot) & - & 26.8 & 30.5 & 32.2 & - & 256M & - \\
\midrule
EgoPulseFormer & EgoPPG-DB & 45.3 & 35.9 & 42.4 & 16 & 121M & 15 \\
SkillFormer & K600 & \underline{45.9} & \textbf{46.3} & \underline{47.5} & 16/24/32 & 14/20/27M & 4 \\
\midrule
\textbf{ProfVLM (MLP)} & K600 + SmolLM2 & 38.9 & 44.2 & 42.9 & 8 & 2.2M & 6 \\
\textbf{ProfVLM (AGP)} & K600 + SmolLM2 & 44.2 & \underline{45.1} & \textbf{48.2} & 8 & 5.3M & 6 \\
\bottomrule
\end{tabular}
\end{adjustbox}
\caption{Comparison with EgoExo4D baselines, with 0-shot general-purpose VLMs \citep{pandagpt, smolvlm}, and with concurrent approaches which adopt the official training-validation splits: EgoPulseFormer \citep{egoppg}, and SkillFormer \citep{skillformer}. Results are reported for Ego, Exos, and Ego+Exos views. Metric is top-1 accuracy (\%) as in the official benchmark \citep{egoexo4d} for direct comparison. ProfVLM+AGP surpasses state-of-the-art accuracy in the Ego+Exos setting using significantly less parameters and input frames. Best in bold, second-best underlined.}
\label{tab:method_comparison}
\end{table}

\begin{table}[t]
\centering
\begin{adjustbox}{width=\textwidth}
\begin{tabular}{lcccccc}
\toprule
\textbf{Method} & \textbf{View} & \textbf{Hid} & \textbf{Heads} & \textbf{Train Params} & \textbf{Acc (\%)} & \textbf{F1 (\%)} \\
\midrule
\multirow{3}{*}{\textbf{ProfVLM (MLP)}} & Ego & 576 & - & 2.2M & 38.9 & 35.4 \\
& Exos & 576 & - & 2.2M & 44.2 & 36.9 \\
& Ego+Exos & 576 & - & 2.2M & 42.9 & 36.0 \\
\midrule
\multirow{3}{*}{\textbf{ProfVLM (AGP)}} & Ego & 1024 & 4 & 5.3M & 44.2 & 38.5 \\
& Exos & 1024 & 4 & 5.3M & \underline{45.1} & \underline{38.9} \\
& Ego+Exos & 1024 & 4 & 5.3M & \textbf{48.2} & \textbf{44.4} \\
\bottomrule
\end{tabular}
\end{adjustbox}
\caption{Accuracy and F1 comparison between ProfVLM variants (MLP vs. AGP) across different view settings. All models share identical hyperparameters: learning rate = 3e-4, LoRA rank = 8, scaling = 32, 8 frames per video, 6 training epochs, batch size = 32. Bold: best results; underlined: second-best.}
\label{tab:profvlm-comparison}
\end{table}

\begin{table}[t]
\centering
% \footnotesize
\begin{adjustbox}{width=\textwidth}
\begin{tabular}{lcccccccccc}
\toprule
& & \multicolumn{3}{c}{\textbf{Baseline}} & \multicolumn{3}{c}{\textbf{ProfVLM (MLP)}} & \multicolumn{3}{c}{\textbf{ProfVLM (AGP)}} \\
\cmidrule(lr){3-5} \cmidrule(lr){6-8} \cmidrule(lr){9-11}
\textbf{Scenario} & \textbf{Maj.} & Ego & Exos & Ego+Exos & Ego & Exos & Ego+Exos & Ego & Exos & Ego+Exos \\
\midrule
Basketball & 36.19 & \underline{51.43} & 52.30 & \textbf{55.24} & 34.00 & 38.00 & 40.00 & 36.00 & 33.00 & 41.00 \\
Cooking & 50.00 & 45.00 & 35.00 & 35.00 & 41.00 & \underline{51.00} & 36.00 & 31.0 & \textbf{56.00} & \underline{51.00} \\
Dancing & 51.61 & \underline{55.65} & 42.74 & 42.74 & 40.02 & 48.97 & 53.03 & 51.41 & 53.85 & \textbf{60.35} \\
Music & 58.97 & 46.15 & \underline{69.23} & 56.41 & 58.89 & 56.26 & 55.26 & \textbf{72.05} & 61.53 & 56.26 \\
Bouldering & 0.00 & 25.31 & 17.28 & 17.28 & 33.08 & \underline{38.11} & 34.33 & 37.48 & 37.48 & \textbf{38.74} \\
Soccer & 62.50 & 56.25 & \underline{75.00} & \underline{75.00} & 69.75 & 69.75 & 44.75 & 57.25 & \textbf{76.00} & 69.75 \\
\bottomrule
\end{tabular}
\end{adjustbox}
\caption{Accuracy (\%) comparison across six scenarios using egocentric (Ego), exocentric (Exos), and combined (Ego+Exos) views. Baseline from~\citep{egoexo4d}. \textit{Maj.} refers to the majority-class baseline. ProfVLM is evaluated with both a simple MLP projector and our AttentiveGateProjector (AGP). Best results are in bold; second-best are underlined.}
\label{tab:scenario-comparison}
\end{table}

\section{Results}
We evaluate ProfVLM on the task of multimodal proficiency estimation using the EgoExo4D \citep{egoexo4d} benchmark. The following sections provide a detailed analysis of our model’s performance: the rationale behind our baseline selection (Sec.\ref{sec:baseline_ratio}), overall accuracy and efficiency (Sec.\ref{sec:overall_acc}), per-scenario accuracy (Sec.\ref{sec:cross-domain}), and qualitative evaluation of the generated textual feedback (Sec.~\ref{sec:language_eval}).

\subsection{Baseline Selection Rationale}
\label{sec:baseline_ratio}
Before presenting our results, we clarify our choice of baselines. We focus our comparison on established multi-view proficiency estimation methods rather than general-purpose vision-language models (VLMs) for several key reasons. First, our approach differs fundamentally from general-purpose VLMs by reformulating proficiency estimation as a unified generation task with specialized multi-view fusion mechanisms—capabilities that general-purpose VLMs inherently lack. Second, adapting general-purpose VLMs to our task would require substantial architectural modifications (e.g., multi-view attention mechanisms, task-specific training protocols), making direct comparison unfair.

To validate this approach, we tested general-purpose vision-language models on the EgoExo4D \citep{egoexo4d} benchmark using zero-shot evaluation with basic frame concatenation (no fusion mechanisms). We chose three representative models with different scales and architectures. First, PandaGPT \citep{pandagpt} serves as our larger-scale baseline—a multimodal model that combines an ImageBind encoder \citep{imagebind} with Vicuna-13B \citep{vicuna} and natively handles video input. Second, we evaluated SmolVLM and SmolVLM2 \citep{smolvlm}, which use a SigLip-93M encoder \citep{siglip} paired with SmolLM2-135M \citep{smollm2}. This second baseline is particularly important because it shares the same language model backbone as our approach, allowing us to assess whether using SmolLM2-135M alone could achieve competitive results without our specialized architectural components.

Results are reported in Table \ref{tab:method_comparison} and confirm that general-purpose VLMs perform significantly worse than our approach due to their lack of proper multi-view fusion mechanisms and task-specific training. Therefore, we focus our evaluation against established multi-view methods like SkillFormer \citep{skillformer} and EgoPulseFormer \citep{egoppg}, which represent the current state-of-the-art for this specific task on the EgoExo4D dataset.

\subsection{Overall Classification Accuracy and Efficiency}
\label{sec:overall_acc}
Table~\ref{tab:method_comparison} presents classification accuracy (\%) across egocentric (1 Ego), exocentric (4 Exos), and combined (1 Ego + 4 Exos) view configurations. Baseline results are taken directly from the EgoExo4D benchmark \citep{egoexo4d}, where models are trained exclusively on video inputs and perform classification using a dedicated output head. In contrast, ProfVLM adopts a vision-language formulation: it receives both visual input and textual supervision, and infers the proficiency level directly within the generated response, without relying on an explicit classification head.

Our best-performing variant, ProfVLM with AttentiveGateProjector (AGP), surpasses the state-of-the-art with 48.2\% accuracy in the combined Ego+Exos setting, outperforming the established EgoExo4D baselines as well as two contemporary works, SkillFormer \citep{skillformer} and EgoPulseFormer \citep{egoppg}, that adopt the same standardized dataset splits. This alignment ensures a fair comparison under identical evaluation protocols. The consistent gains across these benchmarks underscore the robustness of our approach.
Table~\ref{tab:profvlm-comparison} shows that ProfVLM (AGP) consistently outperforms its MLP counterpart across all configurations. The gain is most notable in the Ego+Exos setting, where AGP achieves 48.2\% accuracy and 44.4\% F1, compared to 42.9\% and 36.0\% for the MLP—a relative improvement of 12\% in accuracy and 23\% in F1. This highlights the effectiveness of our structured, attention-based cross-view fusion.

ProfVLM is also considerably more efficient: it requires just 8 frames per video compared to 16–32 in prior approaches, and utilizes only 5.3M trainable parameters versus 14–27M in SkillFormer (5x reduction) and 121M in TimeSFormer-based models (20x reduction). Furthermore, our model converges in just 6 training epochs compared to 15 for TimeSFormer and EgoPulseFormer variants, representing a 60\% reduction in training time. These findings validate our vision-language formulation, which unifies classification and explanation within a compact, interpretable framework.

\subsection{Cross-Domain Accuracy Analysis}
\label{sec:cross-domain}
Table~\ref{tab:scenario-comparison} breaks down performance across six diverse scenarios in the EgoExo4D dataset. Our analysis reveals that ProfVLM outperforms the baselines in five out of six scenarios, with Basketball being the notable exception. Specifically, ProfVLM (AGP) achieves the highest accuracy in Dancing (60.35\%), Bouldering (38.74\%), Cooking (56.00\%), Soccer (76.00\%), and Music (72.05\%), demonstrating remarkable cross-domain effectiveness.

Several interesting patterns emerge from this analysis. First, the benefit of combining views varies by scenario: in Dancing, the Ego+Exos configuration significantly outperforms individual approaches (60.35\% vs. 51.41\%/53.8\%), while in Soccer the exocentric view provides the strongest signal (76.00\%). Second, the structured fusion mechanism in AGP consistently enhances performance over the simpler MLP projector, particularly in visually complex scenarios like Dancing and Bouldering. These domain-specific results highlight ProfVLM's ability to leverage complementary information from multiple viewpoints, where activities requiring spatial relationships and body positioning analysis (Dancing, Bouldering) benefit from integrated egocentric and exocentric information, while activities with standardized visual cues visible from specific angles (Music) perform equally well with single-view configurations.

The Basketball scenario warrants deeper investigation as it represents the only domain where our approach underperforms the baseline (41.00\% vs. 55.24\%). This challenge likely stems from basketball's unique proficiency assessment requirements that differ fundamentally from other activities~\citep{pan2025basket}. Recent research has established basketball as one of the most analytically complex sports, with the BASKET dataset ~\citep{pan2025basket} demonstrating that state-of-the-art video models achieve accuracy barely above random chance despite advances in computer vision. Unlike domains where proficiency manifests through broader movement patterns, basketball skills depend on highly precise technical details: wrist angle during shooting, fingertip control in dribbling, and subtle postural adjustments~\citep{sciencedirect2020sports}. Studies have shown that basketball analysis requires sophisticated spatio-temporal modeling to capture the complex interactions between players, ball dynamics, and tactical formations~\citep{nature2025basketball,tacticexpert}. Our current temporal sampling strategy and spatial resolution may be insufficient to capture these fine-grained discriminative features that distinguish skill levels, particularly given basketball's requirements for spatial accuracy and high temporal resolution~\citep{nature2025basketball}.

Additionally, the attentive fusion mechanism, while effective for integrating complementary information across views in other domains, may struggle with basketball's localized critical regions (hands, ball trajectory) where relevant signals could be suppressed by global attention patterns. The baseline's superior performance suggests that simpler aggregation strategies may be more robust when discriminative features are highly localized and require precise temporal alignment, pointing toward the need for domain-adaptive attention mechanisms and higher-resolution temporal modeling for technical sports analysis in future work.

\subsection{Ablation Studies}
\label{sec:ablations}
We conduct systematic ablations to validate our AGP design choices (Sec. \ref{sec:ablagp}), evaluate the impact of supervision modality and training data volume (Sec. \ref{sec:ablation_supervision}), and examine an alternative prompt formulation (Sec. \ref{sec:ablation_prompt}).

\subsubsection{Projector Architecture}
\label{sec:ablagp}
Table~\ref{tab:projector_ablation} reports the progressive ablation of the AttentiveGatedProjector. Starting from the MLP baseline (42.9\%), we introduce: (1) multi-head attention for cross-view fusion, which corresponds to switching from the MLP to the base AGP; (2) learnable gating for feature selection; and (3) adaptive normalization for vision-language alignment. Each component contributes to the final performance (48.2\%), with the full AGP achieving a 12.3\% percentage increase over the MLP baseline. This validates that all three mechanisms are essential for effective multi-view proficiency estimation.

\begin{table}[t]
\centering
\small
\begin{tabular}{lccccl}
\toprule
Projector & Gating & Adapt. Norm & Params & Acc (\%) & \% $\Delta$ \\
\midrule
MLP & - & - & 2.2M & 42.9 & - \\
AGP (base) & \xmark & \xmark & 4.2M & 44.2 & +3.0\% \\
+ Gating & \cmark & \xmark & 5.3M & 45.5 & +2.9\% \\
+ Gating + Norm & \cmark & \cmark & 5.3M & \textbf{48.2} & +5.8\% \\
\bottomrule
\end{tabular}
\caption{Progressive ablation of AttentiveGatedProjector components using Ego+Exos views. Hyperparameters follow Section~\ref{sec:implementationDet}. Percentage increase (\% $\Delta$) is computed relative to the previous row.}
\label{tab:projector_ablation}
\end{table}

\subsubsection{Supervision Modality and Volume}
\label{sec:ablation_supervision}
We investigate how different supervision strategies affect proficiency estimation performance (Table~\ref{tab:supervision_ablation}).

\paragraph{Label-Only Generation}
We train ProfVLM to generate only the discrete proficiency label (e.g., ``Intermediate Expert'') without commentary, using the official train set (1904 samples). This configuration achieves 45.0\% accuracy, demonstrating that the generative vision-language formulation remains competitive with baselines even when limited to classification-only supervision.

\paragraph{Single Commentary per Video}
Training with exactly one randomly sampled commentary per video yields 40.7\% accuracy, which is below the label-only configuration. This counter-intuitive result reveals an important training dynamic: the commentary generation task introduces loss imbalance. Since loss is computed over all generated tokens, the commentary dominates the gradient compared to the single label token. With only 1904 training samples, the model overfits to generating plausible commentary at the expense of label prediction accuracy. This phenomenon does not occur in the label-only setting (single prediction token) or in the multiple commentary setting where increased data volume and label-commentary diversity provide implicit regularization. 

\paragraph{Multiple Commentaries per Video}
Our full training regime leverages all available expert commentaries per video (4238 samples total, 2.2x data augmentation). This achieves 48.2\% accuracy through: (1) increased training data volume, and (2) implicit regularization: observing the same video with multiple valid commentaries teaches the model that proficiency labels remain invariant to commentary style, strengthening label prediction robustness. The gap between single commentary (40.7\%) and full supervision (48.2\%) demonstrates that effective commentary integration requires both sufficient data volume to overcome loss imbalance and diversity to provide regularization benefits.

\begin{table}[t]
\centering
\small
\begin{tabular}{lc|cc}
\toprule
\textbf{Training Condition} & \textbf{Samples} & \textbf{Acc (\%)} & \textbf{F1 (\%)} \\
\midrule
Label-only generation & 1904 & 45.0 & 36.7 \\
Single random commentary & 1904 & 40.7 & 35.9 \\
Full (all commentaries) & 4238 & \textbf{48.2} & \textbf{44.4} \\
\midrule
\multicolumn{4}{l}{\textit{Discriminative baselines:}} \\
SkillFormer~\cite{skillformer} & 1904 & 47.5 & - \\
EgoPulseFormer~\cite{egoppg} & 1904 & 42.4 & - \\
TimeSformer (best)~\cite{egoexo4d} & 1904 & 40.8 & - \\
\bottomrule
\end{tabular}
\caption{Supervision modality and volume ablation using Ego+Exos views with the AGP projector. Discriminative baselines are reported for reference. Hyperparameters follow Section~\ref{sec:implementationDet}.}
\label{tab:supervision_ablation}
\end{table}

\begin{table}[t]
\centering
\begin{tabular}{lcc}
\toprule
Prompt Structure & Acc (\%) & F1 (\%) \\
\midrule
Proficiency Commentary → Proficiency Level & 45.3 & 41.5 \\
Proficiency Level → Proficiency Commentary & \textbf{48.2} & \textbf{44.4} \\
\bottomrule
\end{tabular}
\caption{Impact of prompt structure using Ego+Exos views. Both variants use the full AGP architecture and multiple commentaries. Hyperparameters follow Section~\ref{sec:implementationDet}.}
\label{tab:prompt_ablation}
\end{table}

\subsubsection{Prompt Structure}
\label{sec:ablation_prompt}
Our default prompt format generates the proficiency label before the commentary (Table~\ref{tab:prompt_format}). To investigate whether this ordering affects performance, we evaluate an alternative format where the commentary is generated first, followed by the label.

Table~\ref{tab:prompt_ablation} shows that generating the label first yields significantly better accuracy (48.2\% vs. 45.3\%). This 2.9\% gap reveals important training dynamics in autoregressive vision-language models. When generating commentary-first, the label prediction at position $T$ depends causally on all preceding tokens at position $\{1, \ldots, T-1\}$ through the attention mechanism. During training, the model conditions on ground-truth commentaries, but at inference must condition on its own generated text. Even when the generated commentary is semantically correct, lexical divergence from training distributions introduces spurious dependencies that harm label prediction.

Additionally, the commentary-first ordering introduces gradient dilution: the label loss must backpropagate through 50-100 commentary tokens, weakening the classification signal with limited training data. In contrast, the label-first ordering establishes the discrete prediction directly from visual features before elaborating with natural language, avoiding spurious causal dependencies in the autoregressive chain. This design choice proves essential for maintaining strong classification performance while enabling interpretable feedback generation.

\begin{table}[t]
\centering
\small
\begin{tabular}{lccc}
\toprule
Model & Epochs & Label Field (\%) & Commentary Field (\%) \\
\midrule
Base (no fine-tuning) & 0 & 0.0 & 0.0 \\
ProfVLM (early) & 1 & 100.0 & 96.7 \\
ProfVLM (final) & 6 & 100.0 & 100.0 \\
\bottomrule
\end{tabular}
\caption{Format compliance results.}
\label{tab:format_ablation}
\end{table}

\subsubsection{Schema Compliance}
\label{sec:ablation_format}
As shown in Table~\ref{tab:prompt_format}, our inference prompt does not provide an output template; therefore, adherence to the expected output structure must be learned during supervised fine-tuning. In Table~\ref{tab:format_ablation} we show how schema compliance emerges during fine-tuning; compliance is measured by verifying both output JSON validity and the presence of the two schema-required keys (``Proficiency Level'' and ``Proficiency Commentary'').
% : (i) 0 epochs (no optimization steps), (ii) 1 epoch, and (iii) full training (6 epochs).

% Format compliance is measured by verifying the presence of the two required fields in the generated output: ``Proficiency Label:'' and ``Proficiency Commentary:''.

% Table~\ref{tab:format_ablation} reports the results. 
It can be seen how the 0-epoch model does not follow the required template. After a single epoch, the model already achieves 100\% compliance for the label field and 96.7\% for the commentary field. At convergence, both fields reach 100\% compliance.
These results indicate that structured output behavior is progressively induced by the supervised training signal and stabilizes with continued optimization.

\subsection{Language Generation Quality}
\label{sec:language_eval}
A key capability of ProfVLM is its ability to generate detailed, expert-like feedback in natural language that explains the reasoning behind proficiency assessments. This dual functionality—providing both classification scores and interpretable commentary—addresses a critical gap in automated skill evaluation where practitioners need actionable insights beyond numerical ratings. We evaluate this language generation capability using established text quality metrics as shown in Table~\ref{tab:vlm_language_metrics}. Since, to the best of our knowledge, no prior work addresses this task, we do not report comparisons with other models in this evaluation.

\begin{table}[t]
\centering
\begin{adjustbox}{width=\textwidth}
\begin{tabular}{lcccc}
\toprule
\textbf{Method} & \textbf{View} & \textbf{BERTScore (F1)} & \textbf{METEOR} & \textbf{ROUGE-L} \\
\midrule
\multirow{3}{*}{\textbf{ProfVLM (MLP)}} 
 & Ego & 85.41 & 18.07 & 15.55 \\
 & Exos & 85.41 & 17.98 & 15.63 \\
 & Ego+Exos & 85.51 & 18.22 & 15.49 \\
\midrule
\multirow{3}{*}{\textbf{ProfVLM (AGP)}} 
 & Ego & 85.41 & 18.06 & 14.47 \\
 & Exos & 85.51 & 17.33 & 15.67 \\
 & Ego+Exos & 85.53 & 18.23 & 15.65 \\
\bottomrule
\end{tabular}
\end{adjustbox}
\caption{Evaluation of generated feedback quality for ProfVLM across different input views and projector architectures. BERTScore (F1) captures semantic similarity to expert commentaries, while METEOR and ROUGE assess lexical overlap. The variations in metric values are minimal across all configurations, suggesting consistent language generation quality regardless of view type or projector design.}
\label{tab:vlm_language_metrics}
\end{table}

\begin{table}[t]
\centering
\small
\begin{adjustbox}{width=\textwidth}
\begin{tabular}{p{9.2cm}rrr}
\toprule
\textbf{Prediction} & \textbf{BERTScore (F1)} & \textbf{METEOR} & \textbf{ROUGE-L}\\
\midrule
\multicolumn{2}{l}{\textit{High-quality example (bouldering):}} \\
So here we can see the climber is going up to their right hand hold. And they have this really nice, pretty strong hold on this left side. But that's not a good choice for this move because it would make them lose power in the same way as they are using it. So the climber was able to kind of get through with this one by just pushing off from the left foot and then putting down into the net so he could come out of his feet. The climbers got out of the grip and now they're hitting this wall. & 87.14 & 36.22 & 20.22\\
\midrule
\multicolumn{2}{l}{\textit{Medium-quality example (dancing):}} \\
So here we have the dancer performing a step up. This is really great because it's so simple and quick, but also very effective. So she does this step up with her arms out to the sides. She's turning around at that point of time. So it feels really nice to see her do this movement in a different direction. Here she steps through and into the space. This is what I like about this move as well. It's kind of like a dance-step where you're stepping forward. & 85.10 & 24.04 & 14.39 \\
\midrule
\multicolumn{2}{l}{\textit{Low-quality example (basketball):}} \\
Again, good job on that. She's playing with her hips and just getting a little bit of power here. But again, you can see she's got a lot more energy in her hands than she should be having. And she has to really focus on the ball right now, which is why we want to know that you're throwing it straight down to your body. Right now, you have to hit this ball up high above your head so that when you release it, it will go off at that point. So I would definitely give this one another shot. & 82.87 & 5.25 & 7.59 \\
\bottomrule
\end{tabular}
\end{adjustbox}
\caption{ProfVLM commentary outputs with corresponding BERTScore (F1), METEOR, and ROUGE-L evaluation metrics. We present three samples representing high, medium, and low quality predicted proficiency feedback.}
\label{tab:generated-text-examples}
\end{table}

\subsubsection{Semantic Alignment and BERTScore Validity}
Our approach achieves consistently high BERTScore \citep{bertscore} F1 values (85.41-85.53) across all configurations, indicating strong semantic alignment with expert commentaries. For BERTScore, we use RoBERTa-large \citep{roberta} as the underlying model, which captures semantic similarity beyond surface-level token matching through contextual embeddings. Importantly, the high BERTScore values reflect genuine semantic coherence rather than mere topical relevance—the metric effectively captures whether the generated feedback addresses the same technical aspects and maintains the domain-appropriate language patterns found in expert annotations. 

These observations also clarify why the AGP and MLP variants yield similar text-generation quality. The language-modeling objective:
\[
\mathcal{L} = -\sum_t \log P(\text{token}_t \mid \text{tokens}_{<t}, \text{video\_features})
\]
creates a strong gradient imbalance (approximately 35:1 between commentary tokens and label tokens), naturally emphasizing fluent and semantically coherent commentary over fine-grained label discrimination. Furthermore, proficiency levels form a continuous spectrum, where expert descriptors often apply to adjacent classes (e.g., \emph{“controlled movement”} may characterize both Intermediate and Late-Expert performances). This structural property of the task enables the model to maintain high semantic alignment even when classification boundaries are mispredicted.

The difference between the two variants manifests instead in visual reasoning: the AGP module integrates multi-view information through cross-view attention and learnable gating, producing more discriminative features and achieving a +12.3\% improvement in prediction accuracy over the MLP. Since both variants share the same language backbone (SmolLM2-135M with LoRA), their commentary fluency remains comparable, while AGP provides stronger visual grounding for proficiency estimation.

\subsubsection{Lexical vs. Semantic Metrics}
The relatively modest METEOR \citep{meteor} (17.33-18.23) and ROUGE-L \citep{rouge} (14.47-15.67) scores reflect the open-ended nature of the feedback generation task. Unlike classification or short-answer generation, our model produces multi-sentence technical commentary where semantic appropriateness is more critical than exact lexical overlap with reference text. This is particularly evident in Table~\ref{tab:generated-text-examples}, which shows that while BERTScore remains relatively stable across quality levels (87.14-82.87), METEOR and ROUGE-L metrics show substantial degradation in lower-quality examples (from 36.22 to 5.25 for METEOR and 20.22 to 7.59 for ROUGE-L). This pattern suggests that ProfVLM preserves domain-appropriate language and conceptual alignment even when diverging from the specific phrasing of expert commentaries.

\subsubsection{Dataset-Driven Feedback Characteristics}
The nature of our generated feedback reflects the characteristics of the EgoExo4D dataset's expert commentaries, which primarily focus on technical observations and skill assessment rather than prescriptive coaching or instructional guidance. Since the original expert commentaries were transcribed from raw audio without manual curation, our model learns to emulate both the expert evaluation style and the naturalistic speech patterns present in the training data, including occasional incomplete sentences and informal language. Our model generates feedback that identifies technical aspects of performance (e.g., "grip positioning could be improved," "timing of the movement shows good coordination") rather than providing detailed corrective instructions. This alignment with the dataset's annotation style explains both the semantic consistency captured by BERTScore and the technical, assessment-oriented nature of the generated text.

\subsubsection{Model Efficiency Considerations}
Our choice of SmolLM2-135M as the language generation backbone represents an explicit trade-off between quality and deployment efficiency. With only 5.3M trainable parameters in the language component, ProfVLM maintains semantic coherence while enabling practical deployment in resource-constrained environments. The consistently high BERTScore values across configurations demonstrate that this efficiency-focused design does not compromise semantic alignment with expert-level technical commentary.

Overall, both quantitative metrics and qualitative analysis confirm that ProfVLM produces coherent and contextually appropriate commentary that maintains semantic alignment with expert annotations while reflecting the technical, observational style characteristic of the underlying dataset.

\section{Limitations}
Despite its promising performance, ProfVLM has some limitations that suggest directions for future work.

First, the model relies on language annotations during training, assuming that both proficiency labels and expert commentaries are available. In real-world scenarios, such supervision may be difficult to obtain at scale or may reflect subjective judgments that vary across annotators.

Second, while ProfVLM demonstrates strong performance in generating coherent and semantically aligned feedback, as confirmed by our automatic evaluation metrics, exploring additional dimensions of quality represents an exciting direction for future work. Although BERTScore effectively captures semantic similarity, aspects such as factual accuracy, pedagogical effectiveness, and actionability could be further assessed only through expert evaluation.

Third, the model generates the class label as part of free-form text. Although this enables seamless integration of classification and explanation, it introduces potential downsides—e.g., misformatted or ambiguous outputs may complicate downstream usage, especially in settings that require structured prediction or reliability guarantees.

Fourth, the current design assumes a fixed number of input views (e.g., Ego and Exos). Adapting the architecture to dynamically handle a variable number of camera perspectives or synchronize inputs with inconsistent frame counts is an open challenge that would improve generalizability.

Lastly, our implementation keeps the video encoder (TimeSFormer) frozen, which simplifies training and reduces computational cost but may limit domain adaptation. Fine-tuning the visual backbone or incorporating domain-specific motion priors could further enhance performance, particularly for tasks requiring fine-grained temporal understanding.

In addition, although ProfVLM leverages multi-view inputs, it primarily relies on high-level visual features, without explicitly modeling pose, body kinematics, or action-specific cues. Explicitly incorporating structured pose and motion representations can potentially provide a more fine-grained understanding of proficiency across views. Investigating how to integrate such structured signals within the vision-language framework represents an important direction for future work.

Addressing these limitations through improved supervision strategies, human-centered evaluation protocols, and more flexible fusion mechanisms represents a promising direction for advancing vision-language models for skill assessment and related multimodal evaluation tasks.

\section{Discussion and Conclusions}
Our evaluation demonstrates that ProfVLM advances proficiency estimation through a unified vision-language framework, surpassing state-of-the-art classification accuracy (48.2\% in Ego+Exos) on the EgoExo4D benchmark while generating semantically aligned textual feedback. The model outperforms video-only baselines and contemporary methods (SkillFormer, EgoPulseFormer) by integrating three key innovations: (1) proficiency inference via language generation instead of explicit classification heads, (2) the AttentiveGateProjector (AGP) for structured cross-view fusion, and (3) a parameter-efficient design requiring only 5.3M trainable parameters and 6 training epochs—yielding up to 20× computational efficiency gains versus transformer-based alternatives.

The AGP mechanism proves critical for multi-view integration, particularly in spatially demanding scenarios (e.g., Dancing, Bouldering), where it outperforms MLP-based fusion by 12–23\% in accuracy and F1. Cross-domain analysis reveals scenario-dependent view utility: exocentric inputs dominate in standardized tasks (e.g., Soccer), while combined views excel in activities requiring spatial reasoning. This adaptability underscores ProfVLM’s capacity to balance complementary perspectives without architectural overcomplication.

Performance in the Basketball domain remains challenging across all approaches, likely reflecting basketball's unique analytical complexity where proficiency depends on highly precise technical details that may exceed our current temporal and spatial resolution capabilities. Addressing these domain-specific challenges represents an important direction for future work.

Beyond classification, ProfVLM generates feedback with strong semantic alignment to expert annotations (BERTScore >85), even when lexical overlap metrics (METEOR, ROUGE-L) are modest—a reflection of its focus on conceptual correctness over rigid template matching. Qualitative analysis confirms coherent explanations across proficiency levels, validating the framework’s dual strength in perception and interpretability.

In summary, ProfVLM establishes a new paradigm unifying classification with natural language reasoning. Its efficient design bridges automated evaluation with expert-like analysis, offering a transparent solution for multimodal proficiency estimation.

\section*{Acknowledgements}
We acknowledge ISCRA for awarding this project access to the LEONARDO supercomputer, owned by the EuroHPC Joint Undertaking, hosted by CINECA (Italy).

\bibliographystyle{elsarticle-harv} 
\bibliography{bib2}
%%       \citet{<label>} ==> Lamport (1994)
%%       \citep{<label>} ==> (Lamport, 1994)
%%
\end{document}